\documentclass[10pt,journal,compsoc]{IEEEtran}
\ifCLASSOPTIONcompsoc
  \usepackage[nocompress]{cite}
\else
  \usepackage{cite}
\fi

\usepackage[T1]{fontenc}
\usepackage[utf8]{inputenc}

\usepackage{enumitem}
\usepackage{subfiles}       
\usepackage{amsmath,xfrac,amsfonts}
\usepackage{multirow}
\usepackage{float}
\usepackage{dblfloatfix}    
\usepackage{bookmark}       
\usepackage{lineno}         
\usepackage{siunitx}        
\usepackage{nth}
\usepackage{url}
\usepackage{longtable}     
\usepackage[ruled,vlined]{algorithm2e}
\usepackage{xcolor}    
\usepackage[flushleft]{threeparttable} 
\usepackage{caption,nameref}
\usepackage{subcaption}
\usepackage{diagbox,makecell}
\usepackage{multirow,diagbox}

\usepackage[numbers]{natbib}
\usepackage{dcolumn,makecell}
\usepackage[utf8]{inputenc}
\usepackage{caption}
\usepackage{pdfpages} 
\usepackage{pgffor} 
\makeatletter
\AtBeginDocument{\let\LS@rot\@undefined}
\makeatother

\def\supplementfilename{supplement.pdf}

\pdfximage{\supplementfilename}
\def\numbersupplementpages{\the\pdflastximagepages}

\newif\ifarXiv
\arXivtrue 

\usepackage{bm, comment} 
\newcommand{\quotes}[1]{``#1''}

\begin{document}



\title{
MINN: Learning the dynamics of differential-algebraic equations and application to battery modeling
}

\author{Yicun~Huang,
        Changfu~Zou,~\IEEEmembership{Senior Member,~IEEE,} 
        Yang~Li,~\IEEEmembership{Senior Member,~IEEE,}
        and~Torsten~Wik,~\IEEEmembership{Member,~IEEE}
\thanks{This work was supported by Marie Sk\l{}odowska-Curie Actions Postdoctoral Fellowships under the Horizon Europe programme (Grant No. 101068764). \\[1.1ex]
All the authors are with the Department of Electrical Engineering, Chalmers University of Technology, Gothenburg, 412 96, Sweden. Emails: yicun@chalmers.se; changfu.zou@chalmers.se; yangli@ieee.org; tw@chalmers.se.
Corresponding authors: Changfu~Zou and Torsten~Wik.}
}

\markboth{}{} 

\IEEEtitleabstractindextext{%
\begin{abstract}
The concept of integrating physics-based and data-driven approaches has become popular for modeling sustainable energy systems. However, the existing literature mainly focuses on the data-driven surrogates generated to replace physics-based models. These models often trade accuracy for speed but lack the generalizability, adaptability, and interpretability inherent in physics-based models, which are often indispensable in modeling real-world dynamic systems for optimization and control purposes.
We propose a novel machine learning architecture, termed model-integrated neural networks (MINN), that can learn the physics-based dynamics of general autonomous or non-autonomous systems consisting of partial differential-algebraic equations (PDAEs).
The obtained architecture systematically solves an unsettled research problem in control-oriented modeling, i.e., how to obtain optimally simplified models that are physically insightful, numerically accurate, and computationally tractable simultaneously. We apply the proposed neural network architecture to model the electrochemical dynamics of lithium-ion batteries and show that MINN is extremely data-efficient to train while being sufficiently generalizable to previously unseen input data, owing to its underlying physical invariants. The MINN battery model has an accuracy comparable to the first principle-based model in predicting both the system outputs and any locally distributed electrochemical behaviors but achieves two orders of magnitude reduction in the solution time.
\end{abstract}

\begin{IEEEkeywords}
Lithium-ion batteries, battery management systems, battery modeling, model simplification, physics-informed machine learning, model-integrated neural networks.
\end{IEEEkeywords}}

\maketitle

\IEEEdisplaynontitleabstractindextext

\IEEEpeerreviewmaketitle

\IEEEraisesectionheading{\section{Introduction}\label{sec:introduction}}

\IEEEPARstart{R}{apid} advances in electromobility have positioned battery as a key player in the transition towards a more sustainable future, with its impact on carbon neutrality continuing to gain momentum. While most battery research has been mainly focused on searching for novel materials~\cite{Materials1,Materials3}, the established battery chain and its circular economy have been dominated by lithium-ion batteries (LIBs) foreseen to prevail due to their proven long-term stability, cost-effective production and recycling. Consequently, the pressure of electromobility has been put on the optimization of LIBs in the foreseeable future, from cell-level chemistry, structure, and manufacturing process to system-level solutions for improved safety, reliability, performance, and lifetime. One key limiting factor in unleashing the full potential of LIBs for electromobility is the current battery management systems (BMS), which limit usage by imposing more or less fixed constraints on battery cell external measurements. The next-generation BMS should enable accurate monitoring and optimal control of dynamical local behaviors distributed inside 
each cell for real-time optimized utilization of the battery systems. 

A battery is a compact, multiphysics system with multiple state variables, domains, material phases and physical parameters over disparate time- and length scales. The current strategies for probing battery internal states involve battery modeling based on equivalent circuits, which at their best, are able to mimic the battery electric behaviors under specific conditions~\cite{Farmann2018ComparativeSO}. More sophisticated electrochemical models for locally distributed internal states, with their minimal assumptions and greater flexibility, offer a distinct advantage over other battery models by providing higher accuracy under a broader range of usage conditions. This makes them an ideal choice for all-purpose and comprehensive battery modeling. Although electrochemical models have been driving a wealth of LIB research in system identification~\cite{Li2020,Khalik2021}, state estimation~\cite{Zou2016,Bizeray2015}, fault and aging predictions~\cite{Yang2017,Jones2022}, and optimal control~\cite{Zou2018,Kolluri2020}, their typical end-user applications, such as smartphones and laptop computers, still require considerable computational power due to the highly nonlinear and stiff PDAEs, let alone the upscaling to pack-level and vehicle fleet-level battery applications. Despite numerous offline implementations employing state-of-the-art numerical techniques~\cite{Torchio2016,Han2021,Berliner2021,Sulzer2021,Korotkin2021}, it is still infeasible to consider advanced electrochemical models for on-board battery management with current hardware.

\begin{figure*}[!hbt]
\centering
   \includegraphics[width=0.75\linewidth]{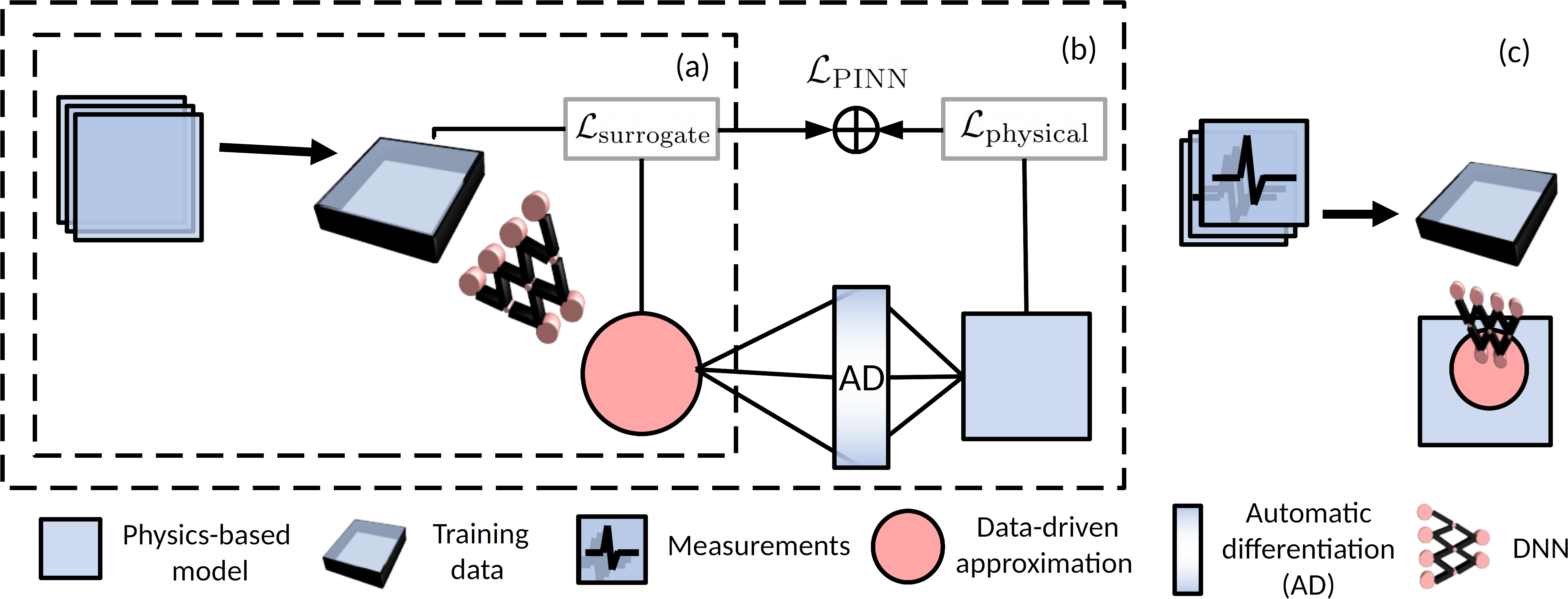}
   \caption{Existing physics-based integration strategies for the blending of neural networks and physics-based models in order to retain their individual merits. (a) A data-driven surrogate model using supervised learning requires relevant and representative training data generated by snapshots of the physics-based model solutions. (b) A surrogate model regularised physical constraints within the PINN framework, of which the PINN loss, $\mathcal{L}_{\text{PINN}}$, is composed of the loss due to model-data inconsistency, $\mathcal{L}_{\text{surrogate}}$, and the loss owing to physical constraints, $\mathcal{L}_{\text{physical}}$. (c) The PINN workflow for inverse problems used to estimate physical parameters as part of parametric PDEs~\cite{Raissi_2017_2}.}
   \label{Fig:PINN}
\end{figure*}

The fundamental challenges in solving PDAEs for control-oriented applications have been commonly addressed by reduced-order modeling. These reduced-order models (ROMs) attempt to lower the computational complexity by either exploring the mathematical structure of the governing equations or simplifying the physics of the original model. For example, 
the single particle model~\cite{SPM} 
features cell-interior variables derived from volume-averaged active materials and uniform molar flux, which result in systems of partial differential equations (PDEs). Other ROMs, such as the physics-based equivalent circuit model~\cite{Li2019,Yang2021}, also result in a simplified system with fewer assumptions. On a high level, these model reduction strategies can be seen as an attempt to replace the original PDAE formulation with a simplified system, resulting in less computational complexity and fewer parameters but compromising accuracy under performance-limiting conditions. 

{\color{black}To circumvent the stiffness issues associated with the first-principle electrochemical models, data-driven approaches have emerged as powerful tools for identifying high-dimensional patterns in battery data. For example, the cycle life of batteries can be accurately predicted using machine learning methods given enough relevant data~\cite{Life1,Life2,Life3}, and fast estimations of the terminal voltage and state of charge (SOC) can be achieved by using recurrent neural networks (RNN)~\cite{LSTM2}.} Nevertheless, training a purely data-driven model to predict internal states of battery cells, such as the electrolyte concentration, local temperature and lithium plating potential, is obscured due to the lack of measurements. However, these internal states are highly important for battery safety, health and performance optimization purposes. Without tracking them, lithium dendrites may ultimately cause internal short-circuits to grow rapidly under conditions of electrolyte depletion~\cite{Harris2009}, high-temperature gradient~\cite{Cui2020} and negative plating potential~\cite{Gao2021}. Notably, as illustrated in Fig.~\ref{Fig:PINN}a, this class of data-driven surrogate models relies on generic neural networks that are agnostic to the underlying dynamics of the battery. Additionally, given a large number of trainable parameters in these neural network models, representative datasets are essential for training to minimize out-of-sample errors, which are often generated by physics-based models. 

\textcolor{black}{In contrast to reduced-order modeling and data-driven approximation, physics-constrained learning offers a distinctive approach to integrate neural networks with physics-based modeling. For instance, neural ordinary differential equations (neural ODEs) leverage the approximation capabilities of neural networks to model dynamic states described by differential equations~\cite{Chen_2018}. Neural ODEs enable a continuous-time formulation of deep learning models, enhancing their flexibility and interpretability in handling dynamic systems. Despite their success in predicting the state of health in batteries~\cite{NODESOH}, the dynamic equations approximated by neural ODEs remain essentially a black box. This limitation prevents the direct integration of domain-specific knowledge, such as current and energy conservation laws or constitutive equations of battery systems, into the learning process. Consequently, it undermines their generalization capability and robustness in predicting the evolution of system-specific states.}

\textcolor{black}{To blend neural networks with governing physical laws directly, the physics-informed neural networks (PINNs) have been introduced to approximate solutions to PDEs by incorporating physical constraints into the loss function~\cite{Raissi_2017_1}. As illustrated in Fig.~\ref{Fig:PINN}b, the PINN framework utilizes automatic differentiation (AD) to compute the residual of PDEs in an unsupervised manner. This process results in a physical constraint loss term, $\mathcal{L}_{\text{physical}}$, which is added to the supervised loss, $\mathcal{L}_{\text{surrogate}}$. 
Additionally, the PINN framework can be employed to estimate the parameters of a physics-based model as schematized in Fig.~\ref{Fig:PINN}c, even with a limited experimental dataset.
Furthermore, PINNs have proven effective even for stiff systems~\cite{Raissi2018}, and various software tools have been developed to automate PINN implementation, making it accessible for different physical systems formulated as initial value problems~\cite{deepxde,SimNet,NPDE}. 
However, PINNs are known to encounter difficulties with complex problems~\cite{Krishnapriyan2021} and are inherently unable to handle non-autonomous systems, or systems with different initial conditions. Specifically, the external control inputs or disturbances of a non-autonomous system, or different initializations of a system's states, will alter the system's dynamical behaviors. 
Thus, employing PINNs to approximate these classes of dynamic systems, such as battery management and control systems, is impractical.
}

To bridge the research gap, this work proposes a model-integrated deep learning framework, termed model-integrated neural networks (MINN), designed to leverage the approximation power of neural networks, and the physical insight and numerical machinery, from those of a physics-based model. MINN is shown to be extremely data-efficient to train and can extrapolate beyond the operating conditions considered in the training data. Furthermore, it retains the physical significance of hidden states and model parameters that can be used directly for system identification, model adaptation, state estimation, and model-based control of LIB. The generality of the proposed framework allows for the easy adoption of other dynamic systems. 



\section{Model-Integrated Neural Networks}
\begin{figure*}[!h]
\centering
   \includegraphics[width=1\linewidth]{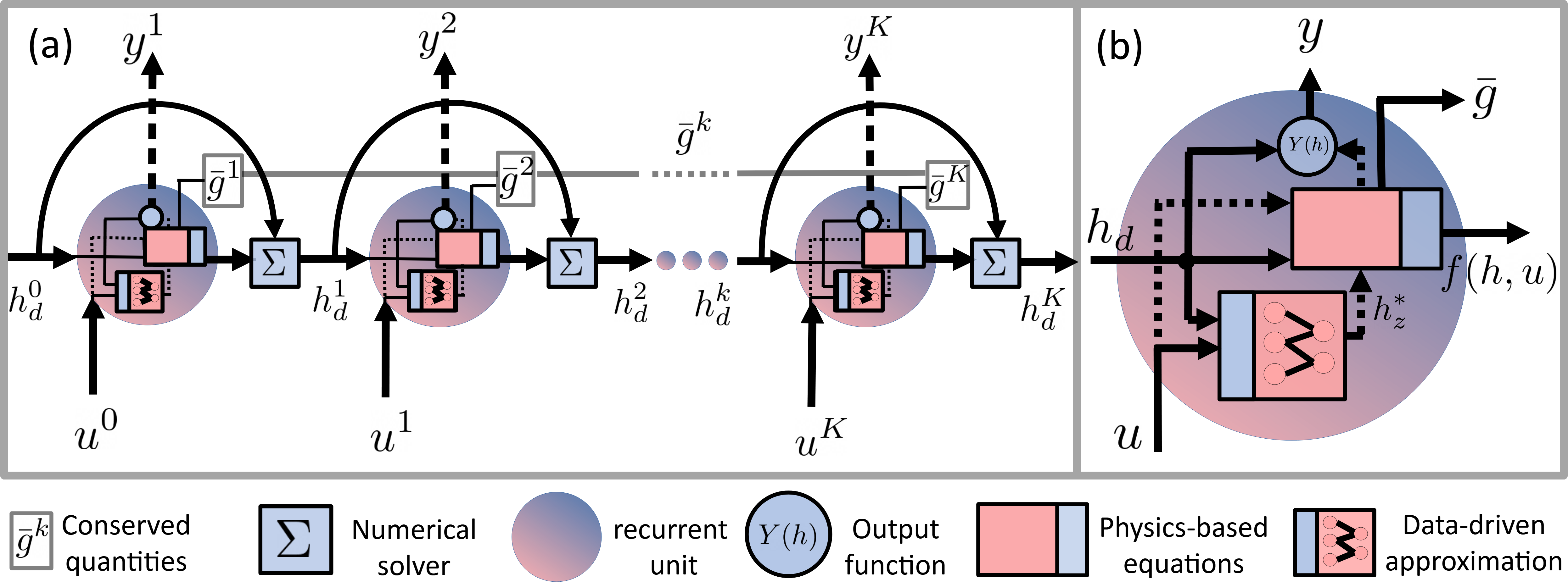}
   \caption{\textcolor{black}{The proposed MINN architecture for dynamic systems. (a) An iterative update of the hidden states $h^k_d$, output $y^k$ and conserved quantities $\bar{g}^k$, is controlled by input $u^k=u(t^k)$ at time $t^k$ through physics-based hidden units. This update is handled by the time integration via the numerical solver. (b) The design of a physics-based recurrent unit contains physics-based equations, a deep learning-enabled approximation ~\eqref{Eqn. DAE08} and an output function $Y$.}}
   \label{Fig:MINN}
\end{figure*}
{\color{black}Different from the two ways of coupling data-driven and physics-based approaches introduced by the PINN framework, this work presents a new hybrid approach leveraging sequence-to-sequence learning to overcome the limitations of PINN. Our approach is designed for a class of generic dynamic systems with control inputs, or different initial conditions. Instead of learning the entire space-time solution using PINN, the main objective of MINN is to derive an explicit function that is implicitly embedded in physics-based models, in order to accelerate the solution process. To achieve this, physics-based equations are integrated into a neural network architecture directly. With the key idea schematized in Fig.~\ref{Fig:MINN}, MINN is formulated and explained step-by-step in this section.}

General multi-timescale dynamic systems can be formulated as coupled differential-algebraic equations (DAEs), resulting from the spatial discretization of the original PDAEs
\begin{align}
\dot{h}_d(t) &= f\Big(t,h_d,h_z,u\Big), \label{Eqn. DAE05} \\ 
y(t) &= Y\Big(t,h_d,h_z,u\Big), \label{Eqn. DAE051} \\
0&=g\Big(t,h_d,h_z,u\Big). \label{Eqn. DAE06}
\end{align}
The above DAE system features differential states $h_d$, algebraic variables $h_z$, and a time-varying control input $u$. $y$ is the output of interest, and $Y$ is the function to compute the output from the states and the input. The origin of the algebraic equation \eqref{Eqn. DAE06} is threefold, i.e., it can stem from the boundary conditions, the singular perturbation of the original PDAEs, or conservation laws naturally arising from the physical problem. In some special cases, explicit solutions for \eqref{Eqn. DAE06} exist for $h_z$, which makes it replaceable by a function $G(t,h_d,u)$ in \eqref{Eqn. DAE05} and \eqref{Eqn. DAE051}. However, in most cases, $h_z$ does not have a fixed-form solution in terms of $t$, $h_d$ and $u$. In such cases, more computationally involved DAE solvers must be used. 

Due to the high computational cost of solving DAE systems, the solution process, or the model itself, must be simplified to suit many model-based applications. If the end result of training PINNs is a fast solution and of numerical methods, a slow, coarse-grained solution, training MINN generates a simplified dynamic model, as shown in the schematics of MINN in Fig.~\ref{Fig:MINN}a. {\color{black}To this end, we parameterize an explicit function $G_{\text{NN}}$ within the recurrent unit by $\theta$. This function can be any nonlinear approximator, e.g. neural networks, as illustrated by the \quotes{data-driven approximation} module in Fig.~\ref{Fig:MINN}b. The colors pink and blue in Fig.~\ref{Fig:MINN} represent approximations and physics-based equations, respectively. For instance, the input of the neural network within the ``data-driven approximation'' module consists of functions of time $t$, differential states $h_d$, and the control input $u$, whereas the output $h_z^{*}$, which is the approximated algebraic variable, is fed to the $g$ function of the \quotes{physics-based equations} module, yielding the approximated conserved quantities $\bar{g}$.  In this manner, the underlying physical invariants and domain-specific priors are encoded in the recurrent unit. }

Mathematically, the hidden states of the proposed MINN framework at time step $k+1$ are updated via the \quotes{numerical solver} module using
\begin{align}
h_d^{k+1} &= h_d^{k} + f\,(t^k,h_d^k,h_z^{*,k},u^k)\cdot\delta t^{k},\label{Eqn. DAE07}\\
h_z^{*,k} &= G_{\text{NN}}\,(t^{k},h_d^{k},\,u^{k};\theta), \label{Eqn. DAE08}
\end{align}
where \eqref{Eqn. DAE07} is a discretized form of the continuous-time dynamic equation \eqref{Eqn. DAE05}, and the algebraic variables $h_z^{*}$ in \eqref{Eqn. DAE08} are approximations to the roots of \eqref{Eqn. DAE06}. The function $G_{\text{NN}}$ offers a shortcut to solving the implicit algebraic equations of the DAE system. Additionally, the time step $\delta t$ taken can be adaptively adjusted by the ODE solver, and the form of  \eqref{Eqn. DAE07}  can vary if implicit or multi-stage schemes are used. The system output $y$ and the conserved quantities of the battery system $\bar{g}$ are computed by 
\begin{align}
y^{k} &= Y(t^k,h_d^k,h_z^{*,k},u^k),\label{Eqn. DAE09}\\
\bar{g}^{k} &= g(t^k,h_d^k,h_z^{*,k},u^k), \label{Eqn. DAE10}
\end{align}
where $\bar{g}$ is an approximation of $g$, computed at each time step $k$. $\bar{g}$ does not strictly vanish due to the approximation error, determined by trainable parameters $\theta$. In addition, the MINN framework integrates the differential equations of the DAE system into the neural network architecture through the physics-based hidden recurrent units, as shown in Fig.~\ref{Fig:MINN}b. 

The search for an optimally simplified model \eqref{Eqn. DAE07}--\eqref{Eqn. DAE09} is cast as the following nonlinear optimization problem 
\begin{align}
\text{arg } \min\limits_{\theta}\: \mathcal{L}_{\text{MINN}}\,(\theta\,;\,\lambda) = \mathcal{L}_{y}(\theta) + \lambda\mathcal{L}_{g}(\theta),\label{Eqn. DAE11}
\end{align}
in which the physics-based recurrent units are used to formulate a physics-constrained loss function with $\mathcal{L}_{y}$ and $\mathcal{L}_{g}$ given by
\begin{align}
&\mathcal{L}_{y}(\theta) = \sum_{k=0}^{K}
\Bigg(Y\Big(t^k,h_d^k,G_{\text{NN}}(t^{k},h_d^{k},\,u^{k}),u^{k}\Big)-\hat{y}^k\Bigg)^2\,\delta t^k,\\
&\mathcal{L}_{g}(\theta) = \sum_{k=0}^{K} \,\Bigg|\,g\,\Big(t^k,h_d^k,\,G_{\text{NN}}(t^{k},h_d^{k},\,u^{k}),\,u^{k}\Big)\Bigg|\,\,\delta t^k, \label{Eqn. DAE12}
\end{align}
where $\hat{y}^k$ are measured outputs, and $K$ is the number of samples in the time time-series. In the training of MINN, we seek parameters $\theta$ by solving for $h_d$ for a given profile $u^k$. The loss function comprises a term $\mathcal{L}_{g}$ to quantify physical inconsistency, which due to conservation laws, is a function of the algebraic variables, plus a loss associated with the (measurable) output of the dynamic system. \textcolor{black}{Unlike methods that require an accurate estimate of time derivatives of the states, such as SINDy~\cite{SINDy}, the loss function of MINN is a sum of model-measurement mismatch and $\mathcal{L}_{g}$. They are obtained by integrating the system via an ODE solver, and this approach is known to be more noise-tolerant~\cite{ODENet}.} During training, the parameterized loss function $\mathcal{L}_{\text{MINN}}\,(\theta)$ is minimized via a gradient-based optimizer, and the Lagrange multiplier $\lambda$ is updated iteratively by the steepest ascent. Through \eqref{Eqn. DAE11}--\eqref{Eqn. DAE12}, it can be seen that the training of $G_{\text{NN}}$ is dependent on the time trajectory of the dynamic system \eqref{Eqn. DAE07}--\eqref{Eqn. DAE10}. This enables the deep integration of neural networks and physics-based models. Once the MINN model is obtained, an array of ODE solvers can be readily employed to reproduce the system's state dynamics controlled by an arbitrary profile $u^k$. 


{\color{black}The architecture of MINN integrates elements from RNN and residual neural networks. RNN, known for its modular and flexible design, is particularly suited for sequence-to-sequence applications. As a bi-directional network, RNN makes a good candidate for modeling dynamic systems with time-stepping of states governed by the system dynamics and control input. The design of RNN’s recurrent units facilitates the integration of physics-based equations, allowing for modifications to neuron connectivity and functionality. Despite its advantages, the baseline RNN architecture is hindered by limitations such as short-term memory and the vanishing gradient problem. These issues can be partially mitigated by incorporating mechanisms like gated recurrent units~\cite{GRU0}, echo state networks~\cite{ESN0} and the long short-term memory~\cite{LSTM0}. Nonetheless, these enhancements do not fully resolve the challenges related to data efficiency, and are still susceptible to overfitting and poor extrapolative capabilities~\cite{Finegan2021}. Therefore, incorporating domain-specific knowledge derived from well-understood systems specified by differential equations is crucial for developing neural network architectures that embed physics-based prior knowledge effectively.}

The MINN model is physically informed of the dynamics of the hidden states $h_d$ by adding skip (residual) connections and input to model the dynamic system controlled by $u^k$. This approach enables the integration of a numerical solver that optimizes the time stepping. \textcolor{black}{During training, the learnable  parameters $\theta$ are updated by backpropagating gradients through the solver. Physically, the input $u^k$ and the hidden states $h_d^k$ at time step $k$ are fed into the recurrent unit, resulting in the approximation of the (algebraic) hidden variables $h_z^k$ used in the calculation of a vector-valued function $f$ representing the time derivative. This facilitates the explicit integration of physics-based equations and meaningful states into the MINN architecture. Different from neural ODE~\cite{Chen_2018}, the model-integrated recurrent units allow for the incorporation of control input, thereby enhancing interpretability and extrapolation capabilities. The recurrent units transform hidden states in a sequence-to-sequence manner, generating a time series of conserved quantities, $\bar{g}^k$, for the physics-constrained loss function.}


\section{Application to battery modeling} 
\begin{figure}[h!]
\centering
   \includegraphics[width=1\linewidth]{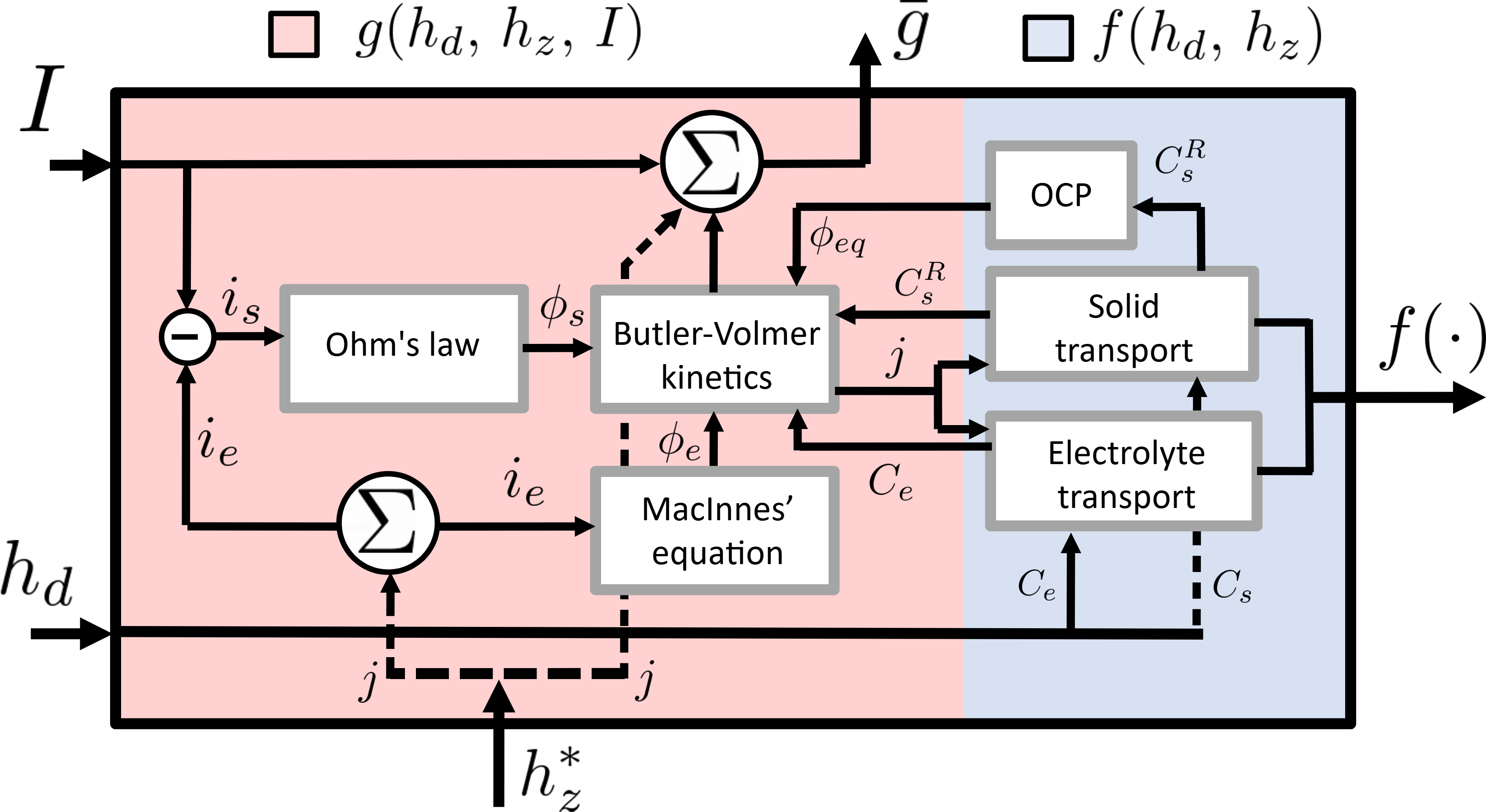}
  \caption{{\color{black}The realization of the \quotes{physics-based equations} module in the physics-based recurrent unit of Fig.~\ref{Fig:MINN}. For battery systems, the control input $u$ is the applied current, i.e. $u^k = I(t^k)$, and the differential state and algebraic variables are represented by $h_d=[C_s,\,C_e]^T$ and $h_z=j$, respectively. The $g$-component evaluates the conservation laws at each time step $k$ with the approximated algebraic variable $h_z^{*}$, while the $f$-component evaluates the time derivative of the differential states $\dot{h}_d$. The two components in the circuitry feature P2D equations, e.g., the open circuit potential (OCP) is a fitted function that takes in the solid concentration at the active material surface and outputs the equilibrium potential $\phi_{eq}$.}}
   \label{Fig:MINN3}
\end{figure}

As stated before, lithium-ion batteries represent a prevalent technology in electromobility and sustainable energy storage that are important forces in the fight against climate change. In this respect, BMS plays a crucial role in battery safety, reliability, sustainability, and dynamic performance. The central thesis for enabling advanced BMS is to develop a battery model that simultaneously preserves physical insights, accuracy and computational efficiency. To this end, the proposed MINN architecture is applied to the modeling of lithium-ion batteries, where each hidden state of the MINN model is assigned to an electrochemical state of the first principle battery model. 
Depending on the application, the control $u$ can be the current $I$ for a battery system. The output may include the terminal voltage, SOC and lithium plating potential if a reference electrode is used. The training data generation for MINN involves only the output that can be measured using, e.g., a three-electrode cell setup. Here, instead of learning blindly from the training data as illustrated by the hybrid approach shown in Fig.~\ref{Fig:PINN}c, we integrate prior knowledge, i.e., the equations from the PDAE system, into the neural network architecture. Fig.~\ref{Fig:MINN3} shows the realization of the physics-based equations in the recurrent unit of MINN, for which the circuitry is based on Newman's P2D model~\cite{Doyle_1993,Newman2,Newman3}. 

\subsection{Physics-based Battery Model}\label{sec:method2}


The most widely used model for Li-ion battery electrochemistry is the celebrated pseudo-two-dimensional (P2D) model, after the paradigm coined by Newman and co-workers~\cite{Doyle_1993,Newman2,Newman3}. The P2D model consists of a set of coupled PDAEs describing the lithium ion dynamics in solid and liquid phases based on porous electrode theory and concentrated solution theory. Although it is a macroscopic model, the model formulations span over multiple length scales. Starting from the pore-scale dynamics within the active particles, the lithium-ion concentration in the solid phase $C_s$ is conserved given a thermodynamic driving force, i.e., the chemical potential $\mu$, according to
\begin{equation}\label{Eqn. solid conservation}
    \frac{\partial C_s}{\partial t}=\nabla\cdot\Bigg( \frac{D_sC_s}{k_BT}\,\nabla\mu\Bigg),
\end{equation}
where $D_s, k_B$ and $T$ are the solid diffusion coefficient, Boltzmann constant and temperature, respectively. The driving force, also termed the chemical potential of the system, adopts the Nernst relation assuming a concentrated solution, i.e., with only entropic contribution $\mu=k_BT\ln C_s$. Consequently, \eqref{Eqn. solid conservation} reduces to Fick's diffusion equation, which in spherical coordinates has the form
\begin{equation}\label{Eqn. Fick}
    \frac{\partial C_s}{\partial t}=\frac{D_s}{r^s}\frac{\partial}{\partial r} \Bigg(r^2\,\frac{\partial C_s}{\partial r}\Bigg),
\end{equation}
where $r$ represents the radial (pseudo) dimension. At the center of the particle ($r=0$), there is a no-flux boundary condition. Imposed by charge transfer, the derivative at the particle surface ($r=R$)  gives the interfacial flux $j$, i.e.,
\begin{equation}\label{Eqn. BV}
    j=D_s\frac{\partial C_s}{\partial r}\Bigg|_{r=R}=\frac{2i_0}{\mathcal{F}}\,\sinh\Bigg(\frac{\mathcal{F}}{2\mathcal{R}T}\eta\Bigg).
\end{equation}

Assuming symmetric Butler-Volmer kinetics, \eqref{Eqn. BV} describes the local reaction molar flux $j$ as a function of (symmetric) exchange current density
\begin{equation}\label{Eqn: i0}
    i_0=\mathcal{F}k_0\sqrt{(C_s^{\max}-C_s^{R})\cdot C_s^{R}\cdot C_e},
\end{equation}
and the local overpotential $\eta$ in the electrode thickness dimension ($x$), i.e.
\begin{equation}
        \eta=\phi_s - \phi_e - \phi_{eq}(C_s^{R}) - R_{\text{SEI}}\cdot j.
\end{equation}

\begin{figure}[h!]
\centering
   \includegraphics[width=0.9\linewidth]{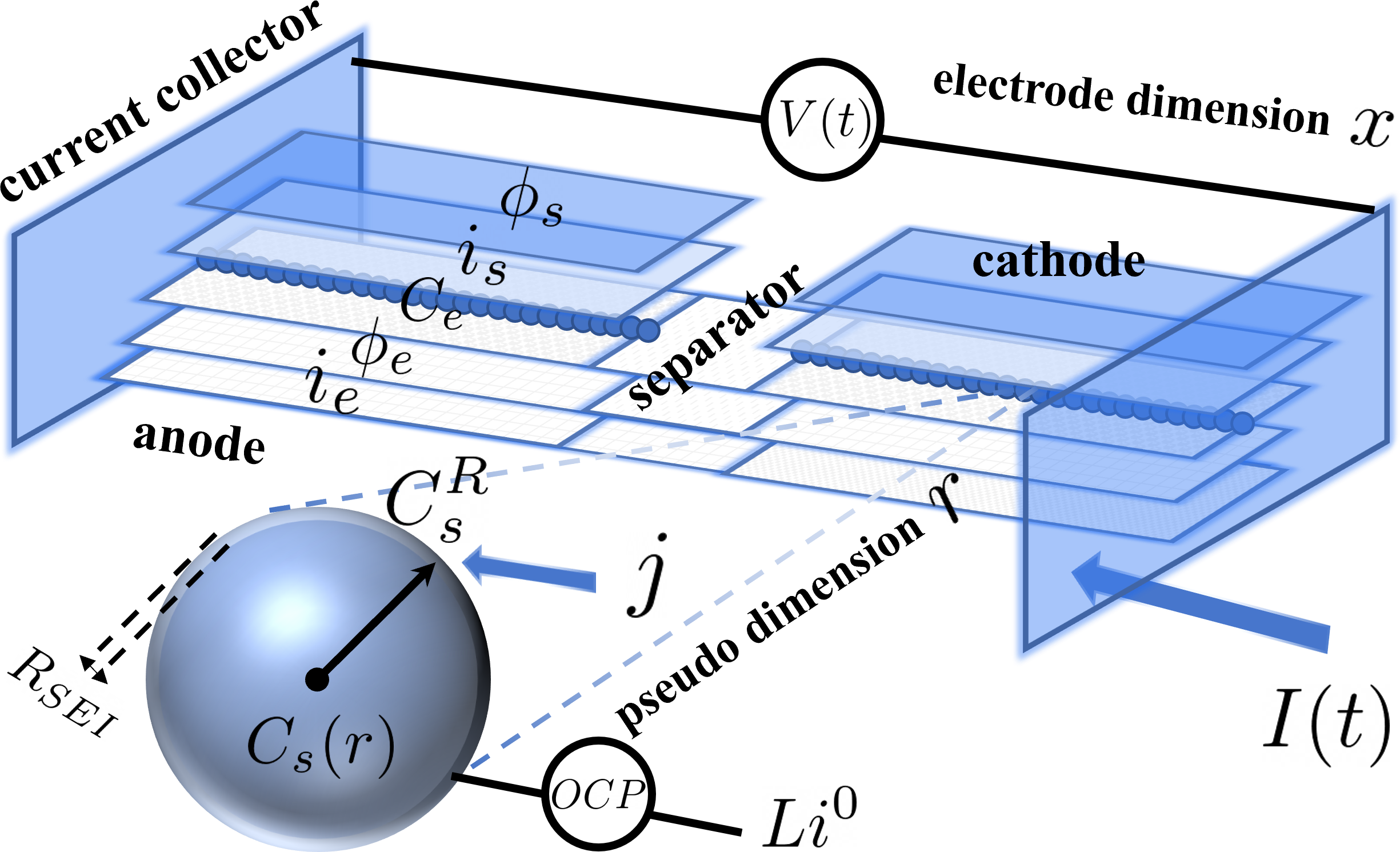}
  \caption{P2D representation of a LIB cell with superimposed continua spanning over two phases and three domains. The nomenclature can be found in the Supplementary Information.}
   \label{Fig:P2D}
\end{figure}

In the above equations, $k_0, C_s^{\max}$, $\mathcal{F}$, $\mathcal{R}$ and $R_{\text{SEI}}$ stand for reaction rate constant, maximum solid concentration, universal gas constant, Faraday constant and resistance of the solid electrolyte interface (SEI). $\phi_{eq}$ denotes the equilibrium potential (relative to lithium metal Li$^0$, determined by the open circuit potential of the materials, which is a function of the concentration at the particle surface $C_s^{R}=C_s|_{r=R}$. $\phi_s$, $\phi_e$ and $C_e$ are the electrical potential, ionic potential and electrolyte concentration fields treated as superimposed continua, along with the currents in the solid and electrolyte ($i_s$ and $i_e$). They are determined by $\phi_s$, $\phi_e$ and $C_e$ according to Ohm's law and the modified Ohm's law, known as MacInnes' equation
\begin{align} 
    \frac{i_s}{\sigma_{\text{eff}}}&=-\frac{\partial\phi_s}{\partial x}, \label{Eqn: Ohm1}
 \\ 
    \frac{i_e}{\kappa_{\text{eff}}(C_e)}&=-\frac{\partial\phi_e}{\partial x}+A\frac{\partial\,\text{ln}\,C_e}{\partial x}, \label{Eqn: Ohm2}
\end{align}
where $\sigma_{\text{eff}}$ and $\kappa_{\text{eff}}$ are effective ionic conductivities in the solid and electrolyte, respectively, and $A=2(1-t_+^0)\mathcal{R}T/\mathcal{F}$ term in the \eqref{Eqn: Ohm2} accounts for the diffusion overpotential induced by an electrolyte concentration gradient. The parallel currents $i_s$ and $i_e$ are constrained by Kirchhoff's law, i.e., $i_s\, +\,i_e\,=\,u(t)$ where $u(t)$ is the applied current. Fig.~\ref{Fig:P2D} illustrates the various fields, domains and boundaries characteristic of the electrochemistry of a Li-ion battery system as per the P2D model.   

To complete the P2D formulation, the electrolyte transport is modeled by a diffusion-reaction equation with a source term that couples it to the lithium-ion diffusion given by \eqref{Eqn. Fick} through the molar, interfacial flux $j$,
\begin{equation}\label{Eqn. eletro}
    \epsilon_e\frac{\partial C_e}{\partial t}=\frac{\partial}{\partial x} \,\Bigg(D_e^{\text{eff}}\,\,\frac{\partial C_e}{\partial x}\Bigg)\,+ \,a_s\,(1-t_+^0)\,j,
\end{equation}
where $\epsilon_e$, $a_s$, $D_e^{\text{eff}}$ and $t_+^0$ are the volume fraction and specific interfacial surface area of the active materials, the effective diffusion coefficient of the electrolyte and the transference number, respectively. The interfacial flux $j$ only exists in the anode and cathode domains and is zero otherwise. 

\subsection{MINN Battery Model}\label{sec:method3}
\noindent Eqns. \eqref{Eqn. Fick}--\eqref{Eqn. eletro} form a system of PDAEs, characterized by the circular, nested loops of algebraic variables $h_z(t)$, which couple the dynamical equations of the differential (dynamic) states $h_d(t)$ through the molar flux terms. The origin of these algebraic variables lies in the enormously different characteristic time scales of ionic and electron transport~\cite{Zou2016AFF}. The resulting system is highly nonlinear and stiff, which poses challenges to numerical techniques. In order to employ specialized solvers optimized for accurate and stable time integration, the system is normally discretized in space, which results in a DAE system in its semi-explicit form,
\begin{align}
\dot{h}_d&=f\left(h_d,\,h_z\right), \label{Eqn. DAE1} \\
0&=g\left(h_d,\,h_z,\,I\right), \label{Eqn. DAE2}
\end{align}
%
%
%
where \eqref{Eqn. DAE1} is an ODE system for ionic transport and \eqref{Eqn. DAE2} are conservation laws resulting from (simplified) electron transport. The so-called hidden states are composed of differential states $h_d=[C_s,\,C_e]^T$ and algebraic variables $h_z=j$. It is also required for the DAE solver to have a consistent initial condition $h(0)=[h_d^0,\,h_z^0]^T $ that satisfies \eqref{Eqn. DAE2}. In the solution process of a DAE system, it must find roots of the algebraic system of equations $g\,(h_d,\,h_z,\,I)$ iteratively within solver tolerances because $h_z$ cannot be explicitly derived. In addition, the system is unstable at an occurrence when the system deviates from $g\,(h_d,\,h_z,\,I)=0$. In such an event, re-initialization is necessary through the discontinuous callback of the DAE solver whenever there is discontinuity detected in the input $I(t)$. Consequently, solving the DAE system originating from the P2D model becomes very expensive in the case of real-world driving cycles, which usually render the DAE solver prohibitively slow. To this end, the proposed MINN model circumvents the need for root finding as well as re-initialization.


The DAE system \eqref{Eqn. Fick}--\eqref{Eqn. eletro} is of index one~\cite{Bizeray2015}, which means that for $h_d$ at a given time $t$, \eqref{Eqn. eletro} defines $h_z$ uniquely. We can therefore find a locally unique solution $h^*_z$ for \eqref{Eqn. DAE2}. Accordingly, the DAE system can be written as one system of ODE,
\begin{equation}\label{Eqn. DAE4}
    \dot{h}_d=f\left(\,h_d,\,\,h^*_z\right).
\end{equation}

The time integration of the ODE system \eqref{Eqn. DAE4} requires no re-initialization and computationally less workload. We then proceed to parameterize the function $G(h_d,\,I)$ by a neural network whose size is dependent on the number of dynamic states (input) and algebraic variables (output). Consider approximating $G$ with a DNN of $L$ layers, i.e., 
\begin{equation}\label{Eqn. NN1}
	\begin{cases}
    a^{[1]} = [h_d,\,I]^T\in \mathbb{R}^{m} \\
    a^{[l]} =  \sigma\,\Big(W_l\,a^{[l-1]}\,+\,b_l\Big),\quad \text{for}\quad l=2,\,3,\,...,\,L \\
    h^*_z = W_La^{[L]}\,+\,b_L \in \mathbb{R}^{n} 
    \end{cases}
\end{equation}
Incidentally, the number of trainable parameters $\theta$ of this approximation $G_{\text{NN}}\,(h_d,I;\theta)$ gets large when the order of the system $N=m+n$ is large, especially if the weight matrices $W_l$, biases $b_l$ and $L$ are also large. An orthogonal collocation method is used for the spatial discretization of the PDAE system \eqref{Eqn. Fick}--\eqref{Eqn. eletro} in this work to relieve the difficulty of training. For the same number of discretization points, this method is known to yield much smaller truncation errors than finite volume~\cite{Torchio2016,Sulzer2021}, finite difference~\cite{Dualfoil} and finite element~\cite{COMSOL,Sulzer2021} commonly used in the battery modeling community, thanks to spectral accuracy. 

There are $16$ boundary conditions in the P2D formulation that are necessary to describe the current and potential fields in three domains. This results in $16$ additional terms signifying the boundary loss in the loss function for a standard PINN setup, which is expensive and difficult to train. The convergence and regularisation of these terms also require additional hyperparameters for tuning. Unlike the physically constrained loss function in the PINN framework~\cite{Raissi_2017_1,Raissi_2017_2}, MINN accounts for the boundary conditions without specifying them explicitly in the loss function. Instead, they are imposed on the integration constants by considering, for example, the ionic potential $\phi_e$ as a function of the electrolyte current $i_e$ integrated over electrode dimension $x$, i.e., 
\begin{equation}\label{Eqn. phie}
    \phi_{e,i} = -\int\frac{i_{e,i}(x)}{\kappa_{\text{eff}}(x)}\,dx + A\,\ln{C_{e,i}\,(x)} + B_i.
\end{equation}
By substituting \eqref{Eqn. phie} into the boundary conditions, three integration constants $B_i$ are obtained for each domain $i$, i.e., anode, separator and cathode. By the same token, the interfacial flux $j_i = \frac{1}{a_{s,i} \mathcal{F}}\frac{\partial i_e}{\partial x}$ can be integrated as follows
\begin{equation}\label{Eqn: ie}
    i_{e,i} = a_{s,i} \mathcal{F}\int j_i\,(x)\,dx + E_i,
\end{equation}
where the integration constant $E_i$ is equal to $I(t)$ in the cathode and $0$ in the anode such that $i_{e}=0$ at the electrode-current collector interfaces and $i_{e}=I(t)$ at the electrode-separator interfaces. In this way, both $i_{e,i}$ and $\phi_{e,i}$ can be exclusively calculated from $j_i$, and so is $i_{s,i}$ because of Kirchhoff's current law, $i_s\, +\,i_e\,=\,I(t)$. Likewise, \eqref{Eqn: Ohm1} can be integrated to yield two integration constants $\phi_{s,i}^{cc}$, which stand for the anode and cathode potentials at the current collector. This gives
\begin{align}
    \phi_{s,i} =\:& \int -\frac{i_s(x)}{\sigma_{\text{eff}}}\,dx + \phi_{s,i}^{cc} \label{Eqn. phis}\\
    \phi_{s,i}^{cc} =\:& \frac{2RT}{\mathcal{F}}\sinh^{-1}\,\Bigg(\frac{\mathcal{F}j_i}{2i_0}\Bigg)+\phi_{e,i} \nonumber \\ & +\phi_{eq,i}-\phi_{s,i}+j_i\cdot R_{\text{SEI}} \\
    Y_V =\:& \phi_{s,a}^{cc} - \phi_{s,c}^{cc}.\label{Eqn: yV}
\end{align}
Eqns. \eqref{Eqn. phie}--\eqref{Eqn. phis} reduce the algebraic variables to only the interfacial flux $j_i$, where $j_i$ is only defined in the anode and cathode. In summary, the algebraic system of equations amounts to 
\begin{equation}\label{Eqn. g}
	g = \begin{bmatrix}
    a_{s,a} \mathcal{F}\int j_a(x)dx  \\
    a_{s,c} \mathcal{F}\int j_c(x)dx  \\
    \end{bmatrix} + \begin{bmatrix}
    -1 \\ 1 
	\end{bmatrix}\cdot I(t). 
\end{equation}
During training, the first term $\mathcal{L}_{y}$ in the loss function \eqref{Eqn. DAE11} measures the error in the model outputs, and they can be the terminal voltage $Y_V$, SOC or lithium plating potential for a battery system. The SOC of the battery system is defined by the average concentration of the anode particles over the electrode thickness $\delta$, normalized by the electrode stoichiometry at $100\%$ and $0\%$ SOC, and the plating potential is the difference between the solid and liquid potentials at the anode-separator interface (ASI), which give
\begin{align}
    &Y_{\text{SOC}}\,(h_d,h^*_z,t) = \frac{3}{\delta\cdot R^3\,(C_s^{100\%}-C_s^{0\%})} \nonumber\\
    & \qquad\quad\:\:\: \cdot \Bigg( \int^{\delta}_0\int^R_0 \frac{r^2}{C_s^{\max}}\,C_s(x,r,t)\,dr\,dx\, -\, C_s^{0\%}\Bigg), \label{Eqn. SOC} \\
    & Y_{\text{plp}}\,(h_d,h^*_z,t) = \phi_s^{ASI}(t)\,-\,\phi_e^{ASI}(t). \label{Eqn: yplp}
\end{align}

\section{Benchmarking and Training}\label{sec:benchmark}
This section introduces four state-of-the-art battery models for evaluating the performance of MINN, using physics-based, data-driven and hybrid approaches. \textcolor{black}{Initially, ground truth solutions are obtained using the P2D model. Subsequently, two benchmarking scenarios are considered: 1) for predefined control input, three baseline designs are developed, including DNN, PINN, and neural ODE battery models; 2) for time-varying control input, which may be unknown {\it a priori} (e.g., derived from real-time optimization or control), a battery model termed data-driven-reduced order model (DD-ROM) is developed to benchmark MINN's performance under real-world driving cycles.}

\subsection{\textcolor{black}{Data Generation}}
Newman's P2D model serves as the ground truth for benchmarking. The model's accuracy is dependent on the spatial discretization of the equations introduced in Section \ref{sec:method2}. To this end, we obtained a high-fidelity P2D model using the spectral collocation method consisting of 130 states and 14 algebraic variables. The resulting 144th-order model is generated symbolically with \textit{Symbolics.jl}~\cite{Symbolics}, and the time integration is done using a legacy IDA solver~\cite{SUNDIALS}. 

\subsection{\textcolor{black}{Baseline Designs}}
\subsubsection{DNN Battery Model}
We developed a purely data-driven battery model using deep learning. This model is parameterized by a three-layer deep neural network \textcolor{black}{with an input size of one, corresponding to the time coordinate, and an output size equal to the number of internal states.} The DNN is trained to map the time coordinates to the corresponding internal states of the battery, $[h_d^k,\,h_z^k]^T$, for a predefined current rate. The loss function is defined as
\begin{equation}\label{Eqn. DNN}
\mathcal{L}_{\text{DNN}} = \sum_{k=0}^{K} \,\Bigg([\mathcal{NN}_d^k,\,\mathcal{NN}_z^k]^T-[h_d^k,\,h_z^k]^T\Bigg)^2\,\delta t^k,
\end{equation}
where $\mathcal{NN}_d^k$ and $\mathcal{NN}_z^k$ are the outputs of the DNN relating to the differential states and algebraic states at timestep $k$, and $\delta t^k$ is the are sampling intervals.

\subsubsection{PINN Battery Model}
The PINN hybrid battery model is developed using the schemes illustrated in Fig.~\ref{Fig:PINN}b. {\color{black}Different from the original PINN formulation~\cite{Raissi_2017_1}, which has a spatial coordinate and time as input, i.e. $[x,\,t]$, the inputs of our PINN battery model are the solution trajectories of all discretized differential states and algebraic variables. Although each of these solution trajectories can be approximated by a PINN mapping space-time coordinates $[x,\,t]$ to seven solution trajectories, i.e. $C_e$ in three domains plus $C_s$ and $j$ in two domains, the training of these PINNs jointly results in a complex loss function that can undermine the performance of PINN, according to \cite{Krishnapriyan2021}. There are also 16 boundary conditions that further complicate the loss function. For these reasons, we formulate the PINN battery model using the discretized P2D model formulation, i.e., \eqref{Eqn. DAE1}--\eqref{Eqn. DAE2}, to include only the time coordinate as the input. In addition to the data-driven loss $\mathcal{L}_{\text{DNN}}$, the loss function for the PINN battery model includes a physical loss due to physical inconsistency:
\begin{align}
\mathcal{L}_{\text{PINN}}=\:&\mathcal{L}_{\text{DNN}} + \mathcal{L}_{\text{physical}},\\
\mathcal{L}_{\text{physical}} =\:& \sum_{k=0}^{K} \,\left( f\left(\mathcal{NN}_d^k,\,\mathcal{NN}_z^k\right)-\dot{h}_d^k \right)^2\,\delta t^k \nonumber \\
&+\sum_{k=0}^{K} \,\Big|\,g\,\left(\mathcal{NN}_d^k,\,\mathcal{NN}_z^k,\,I^{k}\right)\Big|\,\,\delta t^k,
\label{Eqn. PINN}
\end{align}
where $\dot{h}_d^k$ is the time derivative of the differential states in the training data generated by the P2D model. The functions $f$ and $g$ are the right-hand sides of \eqref{Eqn. DAE1} and \eqref{Eqn. DAE2}, respectively, from the P2D formulation.}

\subsubsection{\textcolor{black}{Neural ODE Battery Model}}
\textcolor{black}{Given that both DNN and PINN battery models learn solution trajectories rather than dynamics, they lack adaptability to changes in initial conditions. To address this, a neural ODE (NODE) battery model is incorporated into the baseline design. Unlike the training of NODE~\cite{Chen_2018} using single shooting, multiple shooting is used for training a NODE due to the battery's stiff nonlinear nature. This approach, akin to that detailed in a prior work~\cite{ODENet}, involves fitting multiple successive trajectory segments. To do this for a battery system, we first approximate the dynamics of the battery by a neural network, i.e., $dh/dt\approx \mathcal{NN}(h)$. It should be noted that the hidden state $h$ here does not differentiate between $h_d$ and $h_z$. The loss function is augmented to accommodate shooting constraints using Lagrangian multipliers.}

\subsubsection{DD-ROM Battery Model} 
\textcolor{black}{For benchmarking MINN under dynamic control input, an idealized battery model termed DD-ROM is developed, as the baseline designs above are limited to predefined control input. DD-ROM is a reduced-order model, as it eliminates algebraic variables, thereby reducing the model's complexity. Moreover, it is data-driven, requiring measurements of all internal states $[h_d^k, h_z^k]^T$. In practice, these measurements can only be obtained using next-generation embedded sensors in laboratory settings~\cite{Finegan2021TheAO}.}

The training dataset for DD-ROM consists of all internal state data with the corresponding labels, i.e., a pair of input-label data $\mathcal{X} = [h_d^k,\,I^k]^T$ and $\mathcal{Y} = h_z^k$. The loss function for DD-ROM is defined by the mean square error (MSE) loss of the training data, in a supervised learning fashion. For $49$ pairs of input-label data taken from the high-order P2D solution, a mapping function $\mathcal{X}\mapsto \mathcal{Y}$ is parameterized by a DNN, and the loss is minimized down to machine zero using an optimizer. The downside of this hybrid scheme is that it depends on a vast amount of training data and has to include all battery internal states, of which most are not measurable but can only be obtained by physics-based models. Nevertheless, we train and present such a model with a never-before-seen current profile.

\subsection{Training Details}
The MINN model offers a new path to generating appropriate battery models by seamlessly blending the features of first-principle-based models with neural network architecture, which retrains the individual advantages of physics-based and data-driven approaches. In practice, because certain neural network parameters $\theta$ may lead to unphysical states in \eqref{Eqn. DAE07}--\eqref{Eqn. DAE10}, we rectify the dynamic, algebraic and output functions $f$, $g$ and $Y$ by introducing rectified exponentials during training, e.g. the rectified square root, $\sqrt{x}\,_{\text{ReLU}} = \sqrt{\max(0,x)}$. In addition, we make the electrolyte concentration strictly non-negative. \textcolor{black}{This is done when calculating the model outputs of all four benchmarking models.}

The DNN in the MINN architecture consists of three hidden layers with the same number of nodes as the other models. The gradient of the loss function is computed using forward mode AD as implemented in {\it ForwardDiff.jl} package in Julia~\cite{Revels2016}. We have noticed that although reverse mode approaches are more efficient in most cases, they are not compatible with the solvers used. Gaussian error linear unit activation is used to mitigate training issues such as vanishing gradients usually associated with RNN~\cite{GELU}. All models in the benchmarks are trained with an AdamW optimizer~\cite{ADAMW}. The learning rate is set to 0.001, and the training ends when the loss function flatlines. Generally, a wide-scale separation in the internal states and model output leads to the imbalance of loss function components. We scale the input-output data by characteristic time and internal state scales in order to approximate the widely separated scales in a single DNN. 


\section{Results and Discussion}
This section evaluates the proposed MINN and compares it against the benchmarking models introduced in Section~\ref{sec:benchmark}. The results of the comparative study are characterized by the model prediction accuracy, data efficiency, physical interpretability, and computational cost. It is worth noting that real-world battery applications could be complicated and in a wide range of usage conditions that may never be included in any training dataset. However, a reliable battery model must be able to generalize to out-of-sample usage profiles so that the levels of battery safety and degradation governed by physical states are not under- or overestimated. To test this critical ability of the referred models, we deliberately limit the range of conditions used to generate the training dataset. 


\subsection{\textcolor{black}{Performance under Predefined Control Input}}
\begin{figure*}[h]
\centering
   \includegraphics[width=0.9\linewidth]{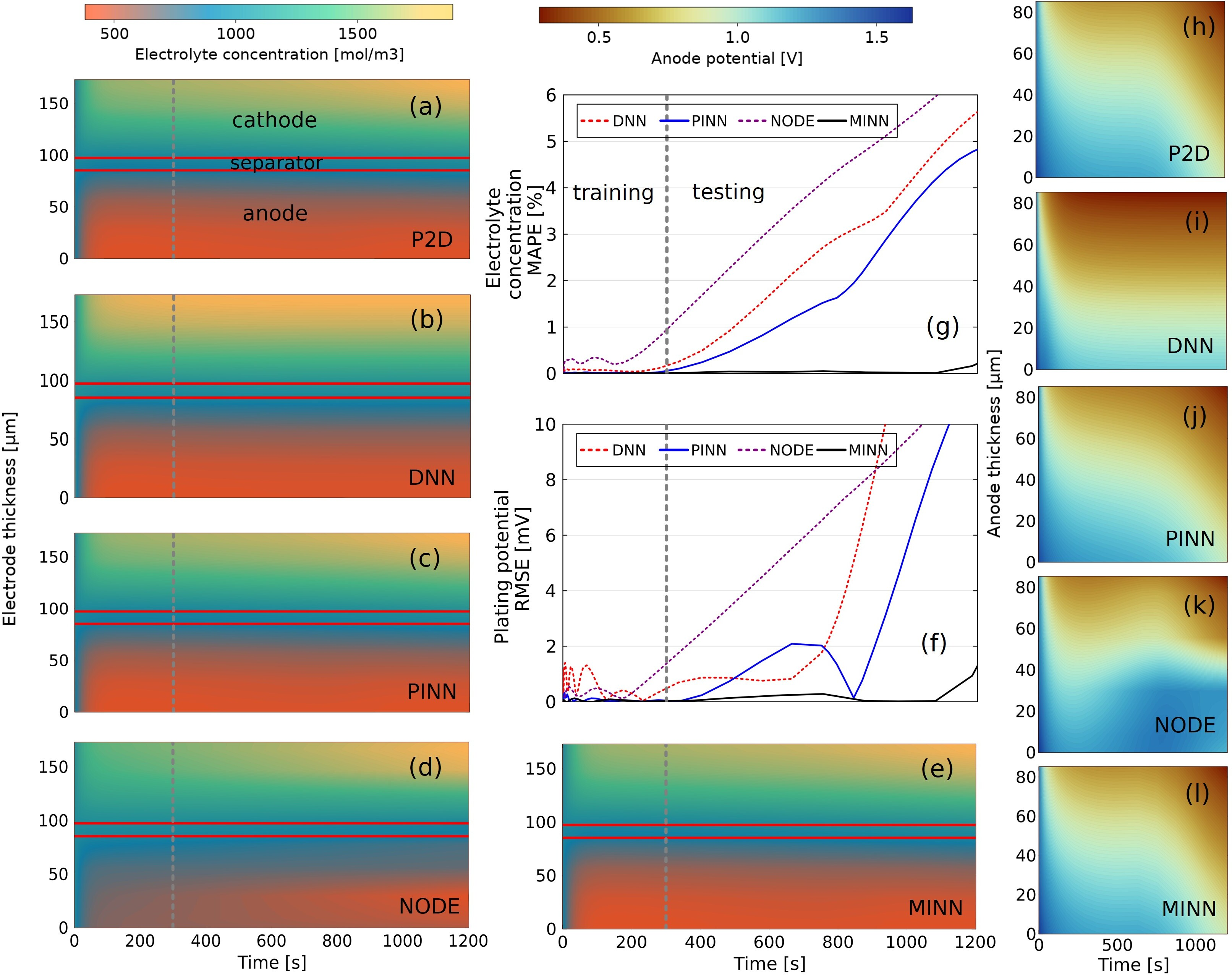}
   \caption{\textcolor{black}{Comparison of different data-driven and hybrid model predictions under 1C charge with P2D predictions as ground truth. All four models are trained using data sampled from the first 300 seconds of the 1C charge data starting from 30\% SOC. 
   (a)--(e) Spatiotemporal plots of electrolyte concentration. (f) The mean absolute percentage error of electrolyte concentration averaged over the thickness. (g) The root mean square error of the plating potential averaged over the thickness. 
   (h)-(l) Spatiotemporal plots of anode potential $\phi_s - \phi_e$.}}
   \label{Fig:pinn}
\end{figure*}

{\color{black} For a predefined control input, such as a constant charge current, the developed DNN, PINN, NODE, and MINN battery models are trained using the first 300 seconds of a P2D spatial-temporal solution under 1C charge starting from 30\% SOC. Testing is conducted from 300 seconds to 1200 seconds. As shown in Fig.~\ref{Fig:pinn}b--e, all four models capture the trend of the time evolution for spatially resolved electrolyte concentration. However, the MAPE of the electrolyte concentration in Fig.~\ref{Fig:pinn}g indicates that the MINN model outperforms the other three models. In particular, the NODE predictions diverge significantly from the P2D reference for the testing data. This divergence may be due to the fact that, although NODE approximates a dynamic system, it does not incorporate domain-specific equations relevant to a battery system. It is well known that electrolyte depletion can lead to safety risks such as lithium dendrite formation and pathological pathways in the battery's aging trajectory~\cite{knees}. Inaccurate predictions of electrolyte concentration will severely undermine the effectiveness of health-aware BMS.

For anode potential, the DNN model also shares the limitation of being model-agnostic. As seen in Fig.~\ref{Fig:pinn}i, the anode potential predicted by the DNN model is unreliable across temporal and spatial coordinates. Both the baseline DNN and NODE models fail to accurately forecast the anode potential trajectories for the testing data. The error in plating potential, shown in Fig.~\ref{Fig:pinn}f, is a critical metric for battery models, as large errors may lead to lithium plating during fast charging applications.

The PINN battery model is capable of following the trend of the spatial-temporal solution, as demonstrated in the electrolyte concentration and anode potential plots. This capability is due to the model's loss function being informed by the P2D equations. The testing error of PINN in Fig.~\ref{Fig:pinn}f  and Fig.~\ref{Fig:pinn}g is due to the fact that the physical loss term \eqref{Eqn. PINN} in the loss function is only a soft constraint~\cite{Krishnapriyan2021}. It is also important to note that DNN and PINN cannot be used for solutions with different initializations. For example, if the initial condition is altered, both DNN and PINN yield completely unreliable results. On the other hand, NODE and MINN benefit from sequence-to-sequence learning, enabling them to account for changes in initial conditions. To illustrate this, we designed an additional test where the initial condition of the testing data differed from that of the training data, the results of which are shown in the Supplementary Information.

The internal state trajectories shown in Fig.~\ref{Fig:pinn} have significant implications for battery health and safety diagnostics and must be accurately captured by the deployed model for next-generation BMS. The MINN battery model consistently captures all local states and model outputs, as illustrated by the spatio-temporal plots in Fig.~\ref{Fig:pinn}e and~\ref{Fig:pinn}l. Furthermore, it can adapt to different initial conditions because MINN learns the dynamics of the physical system rather than the solution trajectories of an autonomous system. Moreover, the MINN model features physically meaningful states and parameters that can be adapted to battery aging.}

\subsection{\textcolor{black}{Performance under Dynamic Control Input}}
\begin{figure*}[h!]
\centering
   \includegraphics[width=0.75\textwidth]{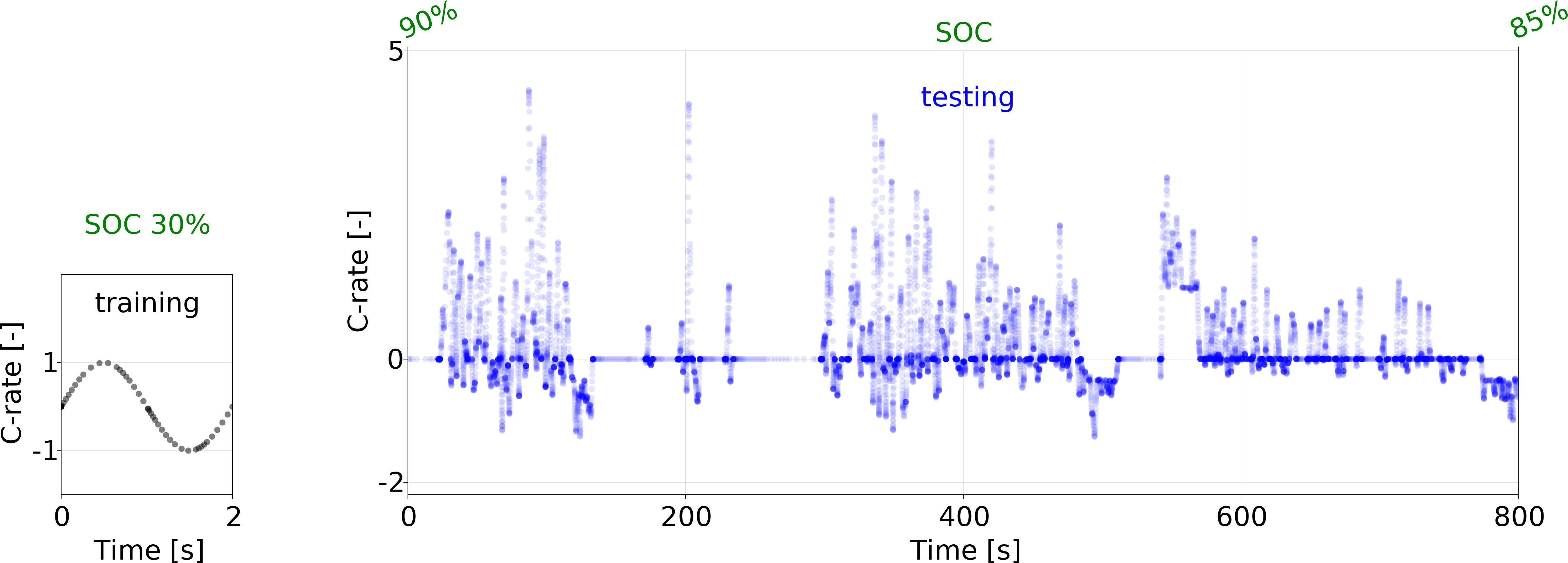}
   \caption{Dynamic current input profile and SOC range for generating training and testing dataset.}
   \label{Fig:It}
\end{figure*}

To evaluate the effectiveness of the MINN model for learning battery dynamics, e.g., excited by a prior unknown input profile, we use an arbitrary vehicle driving cycle to generate testing data. A challenging training dataset is purposefully chosen to evaluate MINN's generalizability on unseen testing data. As shown in Fig.~\ref{Fig:It}, the training dataset consists of 49 snapshots generated by the P2D model with the initial SOC fixed at $30\%$ and a sinusoidal input signal lasting only two seconds and bounded by 1C. By contrast, a highly dynamic current is used in the testing where the initial SOC is set to 90\% and the maximum current reaches 5C. 


For an ideal case where all internal battery states are measurable, the DD-ROM is developed, whereby part of the model states are obtained by a relationship learnt from sampling the state trajectories of the P2D model. In comparison, the training of the MINN model involves only the experimentally measurable output of the P2D model, including the terminal voltage, lithium plating potential and SOC. Thanks to the built-in, problem-specific recurrent unit, it does not require the acquisition of the internal state data. 

\begin{table*}[h!]
\centering\setlength{\extrarowheight}{0pt}
\centering
\begin{tabular}{|*{10}{c|}}
\hline
 \multirowcell{6}{\diagbox[width=9em]{Models}{Metrics} } & \multicolumn{5}{c|}{Complexity} & \multicolumn{1}{c|}{Training} & \multicolumn{3}{c|}{Generalization error} \\
 \cline{2-10}

&\makecell{System} &\makecell{Model\\ order} & \makecell{Solver}&\makecell{\textcolor{black}{Condition}\\ \textcolor{black}{number}} & \makecell{\textcolor{black}{Solution}\\\textcolor{black}{speedup}\\ \textcolor{black}{(average)}} & \makecell{Battery\\ dataset} & \makecell{$\eta_{plp}$ \\~[mV]} & \makecell{Voltage\\~[mV]} & \makecell{SOC\\~[\%]} \\

\cline{1-10}
\hline\hline
P2D & DAE & $130+14$ & \makecell{IDA\\(SUNDIALS)} &\textcolor{black}{$6.0\times 10^{19}$}& \textcolor{black}{1X} &\diagbox[innerwidth=1em, height=\line]{}{} & \diagbox[innerwidth=1em, height=\line]{}{} & \diagbox[innerwidth=1em, height=\line]{}{} & \diagbox[innerwidth=1em, height=\line]{}{} \\
\cline{1-10}
DD-ROM & ODE & $130$ & \makecell{Rodas4} &\textcolor{black}{$6.7\times 10^{9}$}& \textcolor{black}{91X}& \makecell{Internal\\ states $h$} & $9.88$ & $9.87$ & $0.635$ \\
\cline{1-10}
MINN & ODE & $82$ & \makecell{Rodas4} &\textcolor{black}{$3.0\times 10^{8}$}&\textcolor{black}{182X}& \makecell{Measurement\\ $Y$} & $6.28$ & $11.6$ & $0.059$ \\
\hline
\end{tabular}
\caption{\label{tab:benchmark} Model complexity, data efficiency in training. The generalization error of DD-ROM and MINN battery models are compared with results of a dynamic driving profile obtained by a high-fidelity P2D model with the LG M50 parameterization (see {\it Supplementary Information: Model Parameterization}). The generalization error is measured in root mean square error (RMSE) against the P2D benchmark. The solver used for the DAE system representing the 144th-order P2D model is a legacy SUNDIALS solver~\cite{SUNDIALS}, and the stiff solver for DD-ROM and MINN employs a 4th-order A-stable stiffly stable Rosenbrock method (Rodas4). All models are implemented using the {\it DifferentialEquations.jl} package in Julia~\cite{Rackauckas2017}.}
\label{Tab:MINN}
\end{table*}

\begin{figure*}[h!]
\centering
   \includegraphics[width=1.0\linewidth]{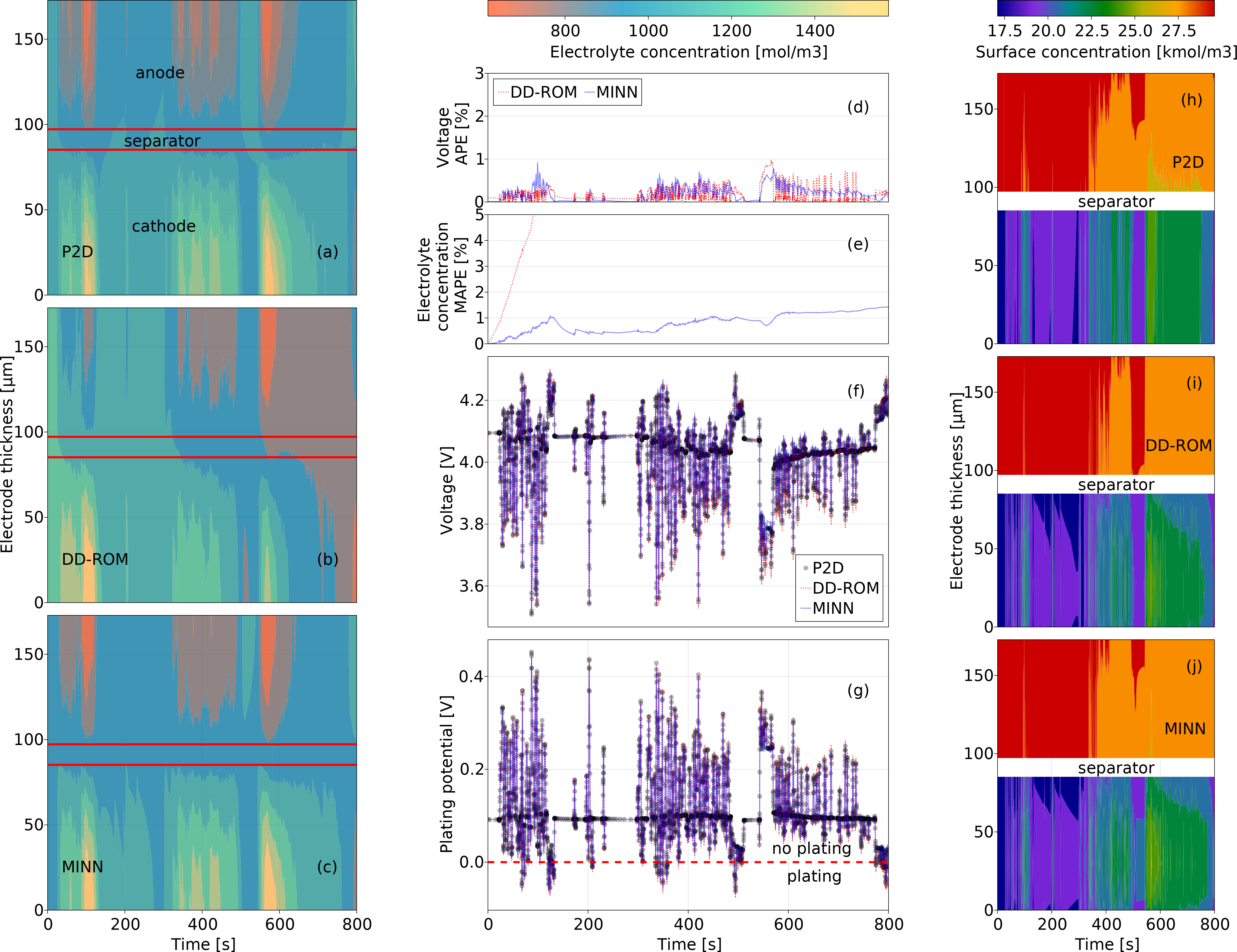}
   \caption{Performance of the DD-ROM and MINN battery models in learning system dynamics, compared with the results of reference 144th-order P2D model. (a)-(c) Liquid-phase ion concentration (electrolyte concentration). (d)-(e) Generalization errors of terminal voltage and electrolyte concentration. (f) Terminal voltage response to the testing current. (g) Anode plating potential at the anode-separator interface. (h)-(j)  Solid-phase concentration at the surface of particles.}
   \label{Fig:MINN2}
\end{figure*}

Accurate prediction of the system outputs, including the terminal voltage, plating potential and SOC, 
is important to advance BMS functionalities, such as power capability prediction \cite{wik2015implementation} and health-aware fast charging \cite{Zou2018,sieg2019fast}. Table~\ref{Tab:MINN} shows the computational complexity, training dataset and numerical accuracy of the above three battery models in predicting these outputs, while Fig.~\ref{Fig:MINN2}g displays the trajectory of the plating potential. Both the MINN model and DD-ROM model show high accuracy in the terminal voltage, 
achieving generalization errors of less than 12 mV. 
In Fig.~\ref{Fig:MINN2}g, the black dashed line at $0$ V highlights the critical level below which the lithium plating is triggered. 
 To predict such plating potential, the MINN model is as good as the DD-ROM which is developed under the hypothetically available information of all battery states. Regarding SOC prediction, the MINN model has greater accuracy than the DD-ROM, with a generalization error of only 0.06\%. 
Other testing data generated at various initial conditions other than 85--90\% SOC have also been considered, which has yielded similar results in predicting the model outputs and further confirmed the superiority of MINN over DD-ROM. This achievement by MINN with only the measurement data for training is practical during real-world battery usage. In fact, the upper limit of the accuracy of MINN is not the first-principle model used in this benchmark but the battery system itself. 




To evaluate MINN's capability of predicting the dynamics of 
internal battery states, 
the locally distributed electrolyte concentration and solid-phase surface concentration are examined in spatiotemporal plots. As shown in Fig.~\ref{Fig:MINN2}c, MINN is capable of reproducing the electrolyte evolution accurately along the challenging operating profile with a MAPE plotted in Fig.~\ref{Fig:MINN2}e of less than 2\%. Although DD-ROM battery model gives less than 1\% absolute percentage error (APE) for the terminal voltage as shown in Fig.~\ref{Fig:MINN2}d, it under- and overestimates the degree of electrolyte depletion. For example, at around 100 seconds, marked by a blue isosurface, 
the DD-ROM model underestimates the depletion in the anode and yet overestimates it in the anode, separator and cathode starting from 600 seconds (light orange). Fig.~\ref{Fig:MINN2}h--j depicts the comparative results for the solid-phase concentration. Here, the colour black represents the theoretical maximum surface concentration of graphite particles, near which the movement of lithium ions matches the intercalation threshold of lattice sites in the anode and the dendritic growth of lithium is inevitable~\cite{Guo2015DirectIS}. The DD-ROM prediction in Fig.~\ref{Fig:MINN2}i yields larger black areas adjacent to the anode-separator interface, which means significantly more lithium plating if one deploys it in the BMS. In addition, DD-ROM overestimates the surface concentration in the anode at 500 seconds. When imposing a constraint on the surface concentration for vehicle battery control, such overestimation will curtail the energy recovered from regenerative braking~\cite{Wikander2019}. However, the MINN model predicts the critical surface concentration in Fig.~\ref{Fig:MINN2}j as close as the P2D during the entire process, thereby allowing accurate monitoring and control of the solid-phase concentration. 




\subsection{\textcolor{black}{Computational Complexity}}
\textcolor{black}{In this section, the computational complexities of DD-ROM and MINN are compared to that of the P2D model. In general, the computational complexity is dependent on the system order. To this end, we define the order of the system as the total number of differential states and algebraic variables. Consequently, the system order is a result of spatial discretization. In our previous work~\cite{Li2022ModelOR}, we evaluated different spatial discretization techniques, e.g. finite volume, finite difference, control volume and spectral methods, for the accuracy of pulse current responses of the solid-phase diffusion. It is found that while most of these techniques capture characteristics in the low-frequency region, some techniques outperform others with the same system order for a wider frequency response range. In particular, the spectral method of order three is more accurate than the 10th-order finite volume method. Although this result cannot be simply extrapolated to the full P2D equations, the assessment of the computational complexity should be based on the DAE or ODE systems derived using the same spatial discretization technique. We have in this regard chosen the spectral method for its superior performance. The resulting system orders of the three battery models in this case study are shown in Table~\ref{tab:benchmark}.}

\textcolor{black}{Additionally, the computational complexity is also characterized by the Jacobian of these battery models. Since timescale separation is determined by the condition number of the Jacobian matrix, we compute the condition numbers at different time steps of the training sinusoidal response. It is equal to the ratio between maximal and minimal eigenvalues of the Jacobian, which roughly measures the local timescale separation. We report in Table~\ref{tab:benchmark} the maximum condition number for the three models. By convention, a system is said to be stiff if this number is large, and it requires stabilized stepping, which is potentially computationally costly.}

In practice, most P2D model implementations feature additional algebraic state variables that are converged at each time step by using, e.g. an iterative algorithm. While these implementations may realize millisecond-scale simulations~\cite{Berliner2021} for static charge and discharge, the solution time can be prohibitively slow for dynamic driving profiles due to increasing stiffness. As shown in Table~\ref{tab:benchmark}, the two hybrid models, i.e., MINN and DD-ROM, achieve two orders of magnitude speedup in the solution time for an 800-second vehicle driving test, of which the MINN battery model has a slight edge over the DD-ROM model. The significant speed improvement of the MINN framework compared to DD-ROM is attributed to its high data efficiency, which allows for learning the complex dynamics of batteries without the need for a fixed number of internal states. This unique feature enables low-order approximations, as demonstrated by developing an 82nd-order model in Table~\ref{tab:benchmark}, in contrast to DD-ROM, which requires 130 states for similar accuracy. The remarkable speedup in computational time will make onboard model-based applications possible, such as online parameter identification, state estimation and closed-loop control. Indeed, a vast majority of daily battery usage is driven by time-varying current profiles, under which the identifiability of battery models, including the P2D and MINN, will often be improved significantly compared to static excitations. 
Therefore, improving computational efficiency under dynamic operating conditions will help lift the computational burden of parameterization.



\subsection{Discussion on Adaptive Battery Modeling} 
During battery lifetime, conventional physics-based modeling requires periodic re-parameterization because of the ever-changing nature of multi-physical battery parameters due to ageing~\cite{Streb2022,Streb2023}. The need for computationally expensive re-parameterization undermines the applications of physics-based models, which are supposed to have minimal dependence on data acquisition and training. This is evidenced by the fact that no mass-produced BMS on the market today has claimed the usage of physics-based models. MINN allows for the simplification of DAE-based structures and may potentially improve the identifiability of parameters. Accordingly, MINN can then be used for aging adaptive models for a wide range of intelligent battery management applications, not only in the short term of several hours or days, but also over the battery's entire lifespan. Numerous examples of such applications include fast charging, lifetime optimization, thermal fault detection, and safety prognosis, which are the main challenges of BMS algorithm design.

\section{Conclusion}
The rapid upscaling of battery-powered electric vehicles makes it possible to collect big data. Based on the data, a wave of data-driven models under the hood of machine learning has recently been developed in the battery community. While data-driven surrogate models excel in learning complex battery characteristics, they inherently lack the ability to generalize beyond the training data and to provide a physical interpretation of the internal battery status. To fundamentally bridge the identified research gap, we proposed a conceptually novel physics-based deep learning architecture, MINN, to seamlessly combine the merits of physics-based and data-driven models. 

MINN stands out through its remarkable accuracy and acceleration, surpassing state-of-the-art benchmarks across the full spectrum of system modeling. In comparison with physics-based simulations, MINN achieved two orders of magnitude speedup while maintaining comparable accuracy. Unlike purely data-driven models, MINN is data-efficient to train and generalizable to unseen operational conditions. Its incorporation of physical parameters and interpretable hidden states, by design, facilitates the learning of system dynamics rather than being limited to input-output relationships. In contrast to the existing practices of physics-based machine learning, MINN offers distinct advantages. Specifically, it can be trained without the need for internal state data, addressing the limitations of current sensing technologies in real-world battery applications. Furthermore, MINN's unique capability to model general non-autonomous PDAE systems under any control input empowers the implementation of advanced and real-time control strategies for internal states. 

The substantial and practical advantages of MINN make it an exceptional choice for developing the next-generation BMS, including battery system identification, fault diagnostics, safety prognostics, and physics-based control. By integrating machine learning with physics-based modeling, the MINN framework offers a powerful tool for analyzing general dynamic systems commonly found in diverse fields, such as mechatronics, thermal fluid dynamics, electrical power systems, and energy storage systems. 


\bibliographystyle{IEEEtranN}
\bibliography{references}

\clearpage
\includepdf[pages=1-7]{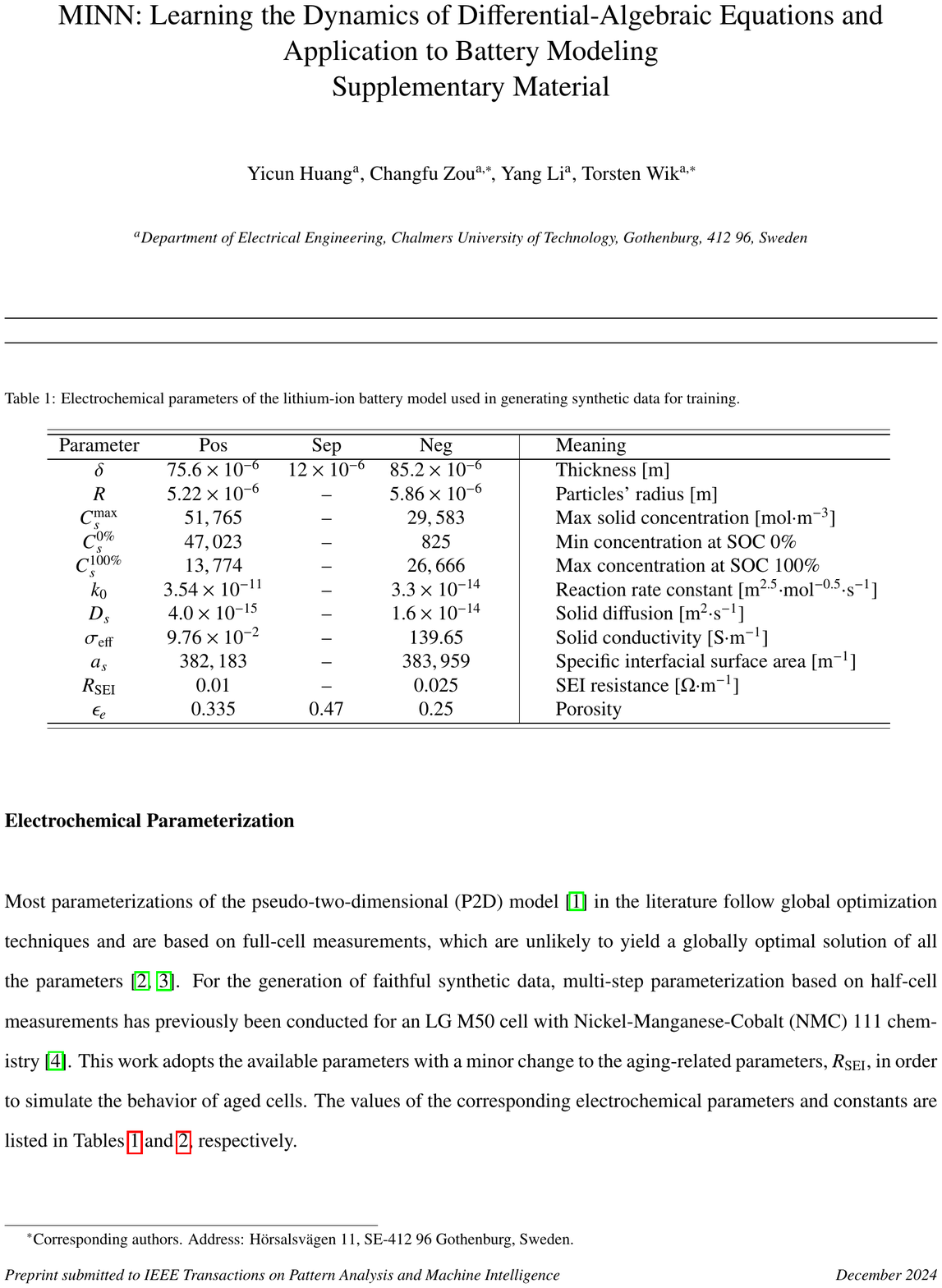}

\end{document}